\pdfoutput=1
\documentclass[11pt]{article}

\usepackage[]{acl}

\usepackage{times}
\usepackage{latexsym}

\usepackage[T1]{fontenc}
\usepackage[utf8]{inputenc}

\usepackage{microtype}

\usepackage{inconsolata}

\usepackage{url}            % simple URL typesetting
\usepackage{booktabs}       % professional-quality tables
\usepackage{amsfonts}       % blackboard math symbols
\usepackage{nicefrac}       % compact symbols for 1/2, etc.
\usepackage{microtype}      % microtypography
\usepackage{setspace}
\usepackage{tabularx}
\usepackage{caption}
\usepackage{amssymb}
\usepackage{multirow}
\usepackage{soul}
\usepackage{url}
\usepackage{algorithm}% http://ctan.org/pkg/algorithms
\usepackage{tcolorbox}
\tcbuselibrary{skins}
\usepackage{xspace}
\usepackage{enumitem}
\usepackage{textcomp}
\usepackage{adjustbox}
\usepackage{wrapfig}
\usepackage{multirow}
\usepackage{amsthm}
\usepackage{subcaption}
\usepackage{comment}
\usepackage[normalem]{ulem}
\usepackage{fancyvrb}
\usepackage{hyperref}   
\usepackage{algpseudocode}
\usepackage{amsmath}
\usepackage{caption}

\algnewcommand{\LeftComment}[1]{\Statex\hspace{\algorithmicindent}\(\triangleright\) #1}

\newcommand{\divse}{\textsc{Div-Se}\xspace}  
\newcommand{\idivse}{\textsc{IDiv-Se}\xspace}  
\newcommand{\auto}{\textsc{DiversePrompting}\xspace}
\title{Diversity of Thought Improves Reasoning Abilities of LLMs}
\author{Ranjita Naik\footnotemark[2]\\
  Microsoft \\
  \And
   Varun Chandrasekaran \\
  University of Illinois Urbana-Champaign \\
  \AND
   Mert Yuksekgonul \\
  Stanford University  \\
  \And
  \textbf{Hamid Palangi}  \\
  Microsoft Research \\
  \And
  \textbf{Besmira Nushi}\footnotemark[2] \\
  Microsoft Research \\
}

\begin{document}
\maketitle
\begin{abstract}

Large language models (LLMs) are documented to struggle in settings that require complex reasoning. Nevertheless, instructing the model to break down the problem into smaller reasoning steps, or ensembling various generations through modifying decoding steps boosts performance. However, these methods assume that the input prompt is fixed and expect the decoding strategies to introduce the diversity needed for ensembling.
In this work, we discuss how one can create and leverage variations of the input prompt as a means of \emph{diversity of thought}. %to improve model performance. 
We propose a method that automatically improves prompt diversity by soliciting feedback from the LLM to ideate approaches that are apt for the problem. We then ensemble the diverse prompts in our method \divse (DIVerse reasoning path Self-Ensemble) across multiple inference calls, or use diverse approaches within a single inference call; we call the latter \idivse (In-call DIVerse reasoning path Self-Ensemble). 
Apart from our approaches outperforming prior work, \textsc{Div-Se} (in particular) advances state-of-the-art performance on the challenging planning and graph  coloring benchmarks. Our results improve the Pareto frontier of the accuracy-cost trade-off.
\end{abstract}  
\footnotetext[2]{Correspondence to \texttt{ranjitan@microsoft.com} and  \texttt{besmira.nushi@microsoft.com}.
}
\section{Introduction}
\label{sec:intro}
% P1: Prompting is important, tasks are hard, but techniques are ad-hoc

\begin{figure*}[t]
    \centering
    \includegraphics[width=0.95\textwidth]{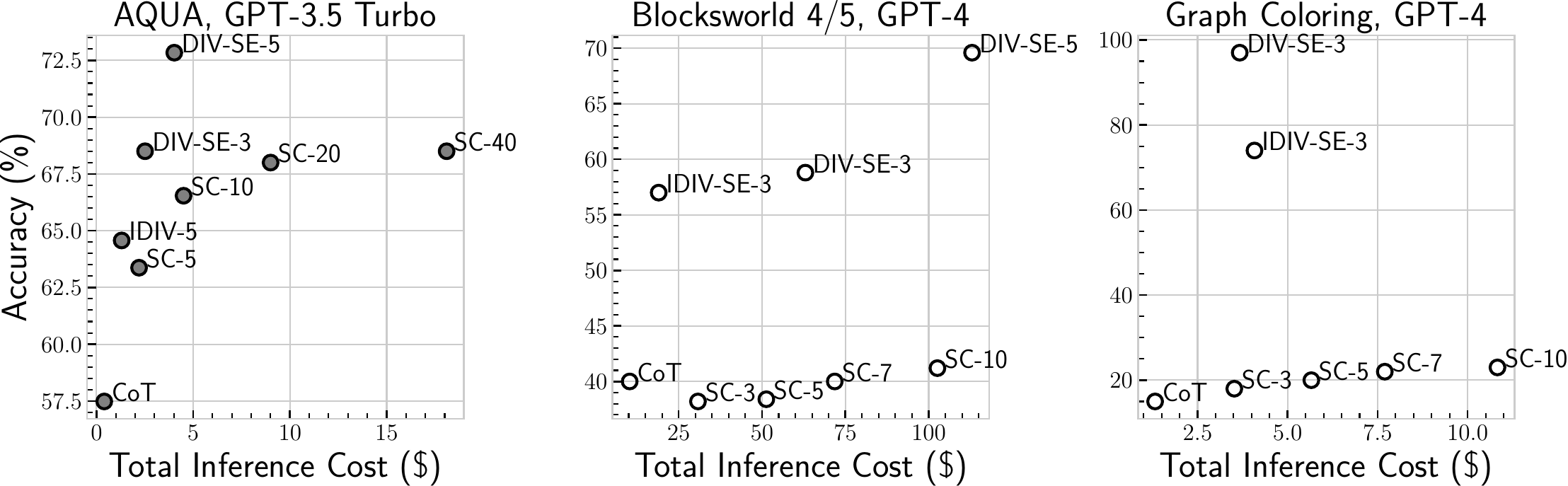}
    \caption{\textbf{Diversity of Thought enhances the inference cost vs. accuracy trade-off.} We compare DIV-SE and IDIV-SE with SC~\citep{wang2023selfconsistency} and CoT~\citep{wei2022chain} across three benchmarks. The x-axis indicates the total inference cost (as defined in \S~\ref{sec:experiments}) on the benchmark using the given method, while the y-axis represents the LLM’s performance. The few-shot-CoT setting is represented by filled gray dots, while the zero-shot-CoT setting is indicated by unfilled dots. Notice that for a fixed cost, our approaches always give better performance.
}
    \label{fig:cost_vs_accuracy_plot_mert}
\end{figure*}

Large language models (LLMs) exhibit state-of-the-art performance across a myriad of tasks, but their effectiveness is strongly influenced by prompt design~\citep{anil2023palm,openai2023gpt4,nori2023can}. For complex reasoning tasks, the right prompt can enable LLMs to capitalize on task structure~\citep{Guidance:2024:Online}, such as by facilitating memory (by externalizing thought processes), or through tractable problem decomposition~\citep{zhou2024self}. However, existing prompt design either relies on iterative trial-and-error~\citep{white2023prompt}, or is expensive~\citep{lester2021power}.

% P2: 2 general principles that are effective.
Previous works identified two simple, yet general prompting principles to enable complex reasoning: (i) Chain-of-Thought~(CoT) prompting, and (ii) ensembling multiple solutions from diverse decoding paths. CoT prompting~\citep{wei2022chain} improves performance by guiding the LLM to follow step-by-step reasoning. Self-consistency~(SC)~\citep{wang2023selfconsistency} instead increases the stochasticity by modifying the decoding process and obtaining multiple completions, which are then ensembled.

% P3: But combining the two is exxpensive $$$$
However, combining the two principles raises limitations. First, inference is significantly more expensive due to numerous runs, each generating long completions with many reasoning steps. Next, it may be impermissible to modify the decoding process in some settings, such as commercial deployments. Finally, stochasticity-based methods do not directly guide the diversity at the level of thought or method, but rather at the token level. This poses limitations because linguistic token diversity does not always ensure diverse and independent solution approaches. 

In this paper, we explore how to explicitly promote the \emph{diversity of thought} while mitigating the aforementioned issues. Prior work by~\citet{li2023making} highlights the importance of prompt diversity, but their notion of diversity is captured through variety in the few-shot examples provided with the prompt; ours focuses on the reasoning approach. We first solicit the LLM to produce multiple-high-level reasoning approaches for problem-solving (e.g., \texttt{method of elimination, visualization techniques} etc. for math reasoning problems). We then leverage GPT-4 to augment few-shot examples used in prior work~\citep{wei2022chain} into the corresponding approaches, whenever applicable.

% Divse, promotes diversity, gets us better perf.
We propose \divse (DIVerse reasoning path Self-Ensemble) to extract and aggregate responses (via majority vote) across multiple inference calls (\S~\ref{subsec:techniques}). Since distinct approaches introduce diversity at the ``thought'' level, our methodology results in improved ensemble accuracy. In Fig.~\ref{fig:cost_vs_accuracy_plot_mert}, we show that it yields more accurate results across multiple reasoning benchmarks at a fixed inference cost, without modifying the decoding procedure. 
For instance, in the \textsc{Blocksworld 4/5} task~\citep{valmeekam2022large}, \divse improves the performance by 29.6 percentage points (p.p). However, this method still leverages multiple inference calls, which could be costly.

% IDIVse, combining the demos.
To reduce inference costs, we build on the observation that the approaches are often mutually independent, and can be combined in a \emph{single prompt} to solicit multiple solutions~\citep{cheng2023batch}. Based on this premise, we propose \idivse (In-call DIVerse reasoning path Self-Ensemble; \S~\ref{subsec:techniques}), which combines all approaches within the same prompt and aggregates all resulting outputs to leverage diversity with a reduced cost. 
Fig.~\ref{fig:cost_vs_accuracy_plot_mert} demonstrates that this method obtains comparable accuracy to \divse and better performance than prior work with lower inference costs. 

% We push the pareto frontier.
We push the pareto frontier of the cost-accuracy trade-off of prompting strategies across multiple reasoning tasks~(\S~\ref{sec:results}), outperforming both CoT and SC prompting on both GPT-3.5 and GPT-4. This is evident from Fig.~\ref{fig:cost_vs_accuracy_plot_mert} for the \textsc{AQuA-RAT} ~\citep{ling-etal-2017-program}, planning ~\citep{valmeekam2023planning}, and graph coloring~\citep{stechly2023gpt4} benchmarks, where there is a performance improvement of 16.52, 29.6, and 82.5 p.p respectively. These improvements, some of which are state-of-the-art, show the potential of thought diversity to extract complex reasoning abilities from LLMs that were impossible to leverage otherwise. We will open source our code upon publication to encourage further research. 
\begin{figure*}[t]
\centering
\begin{minipage}[t]{0.9\textwidth}
\includegraphics[width=\linewidth]{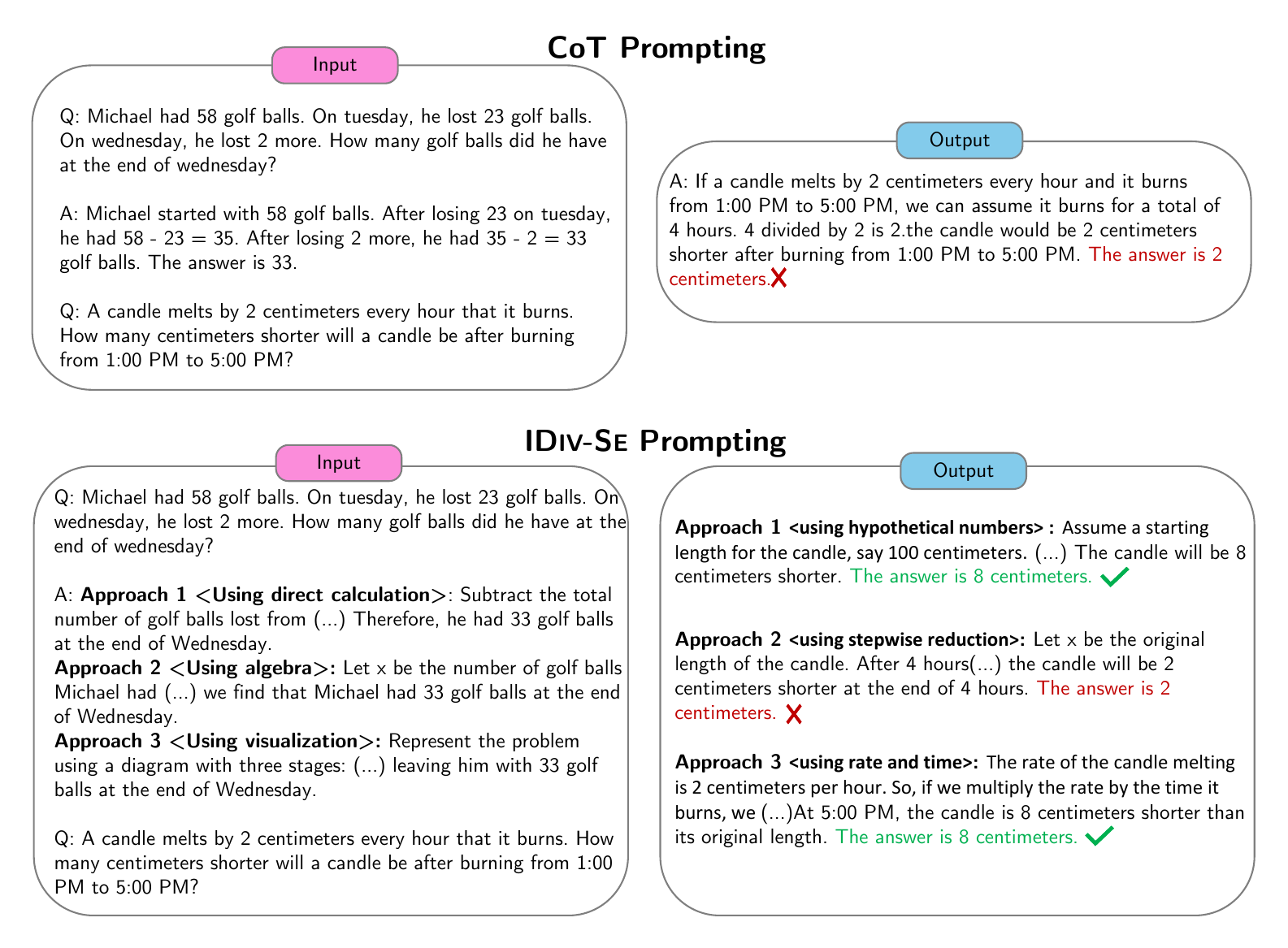}
\end{minipage}
\caption{\textbf{Diversity of Thought}. This illustration depicts CoT and \idivse prompting strategies. Notice that both have a single example. However, \idivse presents more diversity in terms of reasoning paths. This enables it to generate diverse completions, yielding more accurate responses.\vspace{-5mm}}
\label{fig:overall}
\end{figure*}

%\section{Soliciting Diversity through LLM Interactions}
\section{Diversity through LLM Interactions}
\label{sec:method}

% \newpage

First, we introduce terms and notations that we use throughout the paper. We use upper case for sets, lower case for variables, and $[n] = \{1, \cdots, n\}$.

% \begin{itemize}
% \itemsep0em
\noindent{\bf Approach:} These are reasoning strategies for problem solving, denoted with the variable $a$. For example, for the \textsc{GSM8K}~\citep{cobbe2021training}, a benchmark of grade-school math problems
, some of the (generated) approaches can include $a_1=$\texttt{``using visualizations''}, $a_2=$\texttt{``working backwards''}, $a_3=$\texttt{``using direct calculation''}, and $a_4=$\texttt{``method of elimination''}.

\noindent{\bf Persona:} In addition to specifying ``how'' to solve a reasoning problem, specifying a persona can also influence how the LLM behaves~\citep{salewski2023context}. %. 
We denote this with the variable $p$ e.g., $p_1=$\texttt{``Thinking like Alan Turing''}, $p_2=$\texttt{``Thinking like Math Professor''} for the \textsc{GSM8K} task. Note that both approaches and personas are dependent on the reasoning problem.

\begin{algorithm}[t]
\footnotesize
\captionsetup{labelformat=empty}
\caption{{\em \auto}: Prompt creation.}
\begin{algorithmic}
\Procedure{DiversePrompt}{\texttt{size, type}, $F, D, V$}
\State \LeftComment \textbf{Step 1: Identify different approaches to be used.}
\State $A = \{a_1, \ldots, a_{n}\} \gets \texttt{det\_approaches}(\texttt{D})$ 
\LeftComment{where $A$ is the set of approaches}
\State \LeftComment \textbf{Step 2: Identify different personas to be used.}
\State $P = \{\phi, p_1, \ldots, p_{m}\} \gets \texttt{det\_personas}(\texttt{D})$ 
\LeftComment{where $P$ is the set of personas}
\State \LeftComment \textbf{Step 3: Find the best combination.}
%of approaches and personas.}
\State $ S = \{s_1, \ldots\, s_{\texttt{size}}\} \gets \texttt{combine}(A, P, \texttt{size}, V)$ 
\LeftComment{where $S$ is the set of combined approaches and personas, and $s_i = (p, a_{i} \in A)$}
\State \LeftComment \textbf{Step 4: Augment the few-shot examples.}
%\State $T = \{f((p_i, a_j), FS), f((p_i, a_k),_FS),\ldots\} \gets \texttt{transfer}(S, F)$ \LeftComment{where $T$ is the set of style-transferred examples}
\State $T = \{\tilde{T}_{i,j}, \ldots\} \gets \texttt{augment}(S, F)$ \LeftComment{where $T$ is the set of augmented examples, and $\tilde{T}_{i,j}$ is formed using $s_i \in S$ and $f_j \in F$; $|T|=$\texttt{size}}
\State \LeftComment \textbf{Step 5: Compose the final prompt.}
%based on $T$, $S$, and \texttt{type}.}
\State $O \gets \texttt{compose}(T, S, \texttt{type})$ 
\State \textbf{return } $O$ \LeftComment{Return the final output.}
\EndProcedure
\end{algorithmic}
\label{alg_1}
\end{algorithm}

\subsection{Using the LLM as a guide}
\label{subsec:ap}

Proposed method for creating prompts, which we term {\em \auto} 
 is presented in Algorithm~\ref{alg_1}.
Below, we will describe each step in more detail. At a high-level, we solicit feedback from the LLM on how to solve tasks.

\noindent{\bf {Step 1+2. Extracting Approaches \& Personas:}} Note that LLMs trained on internet-scale data encode a significant knowledge from multiple domains~\citep{liang2022holistic,bubeck2023sparks}. While LLMs may not be perfect at solving reasoning tasks, we hypothesize that they are helpful in providing high-quality intermediate feedback.

To extract approaches, we utilize the following methodology: (i) Randomly picking a question from the reasoning dataset $D$ we want to evaluate; and (ii) Creating an instruction prompt where we ask the LLM to generate the names of $b \in [1,5]$ {\em approaches} to solve the aforementioned question conforming to a predefined template (for easier post-processing). Refer to Figure~\ref{fig:auto-example} for an example of the prompt used. 

We extract the part of the response that is compliant with the template and store it. We repeat this process $c$ times (obtaining of $c \cdot b$ candidate approaches), and pick the $n$ most frequent approaches to store in set $A$\footnote{\label{note1}In practice, we set $c=100$, $b=5$, $n \in \{3, 5\}$, and $|V| < 20$.}. This process is abstracted as method \texttt{det\_approaches(.)}.

One can repeat the above process used to extract relevant personas for a given reasoning task. However, we followed a simpler route and asked the model directly for relevant personas for a given task and then included them in the set of $m$ candidate personas $P$ used. This is abstracted as method \texttt{det\_personas(.)}. Note that no persona ($\phi$) is also part of the persona set.

\noindent{\bf {Step 3. Choosing the Best Persona, Approach Pairs:}} The choice of persona and approaches introduces a principled way to promote diversity. 

If the set of personas is $P$, and the set of approaches is $A$, the Cartesian product of $P$ and $A$ yields the total number of prompts. In practice, for each combination (denoted by $s_i$) of persona and approach, we evaluate the prompt formed using the composition on a small validation set $V$\textsuperscript{\ref{note1}} and choose the best performing ``\texttt{size}'' elements on the given task\footnote{For a given reasoning task, we perform this process once~(for GPT-3.5 Turbo), and re-use our selection across all LLMs we evaluate. }.

\noindent{\bf {Step 4. Augmenting few-shot examples:}} Once the (subset of) approach and persona pairs are fixed, we ask the LLM to augment existing few-shot examples
(denoted $F=\{f_1, \cdots\}$) with the given set of approaches. Specifically, we take the few-shot examples provided by~\citet{wei2022chain}, and ask the LLM to solve them in the style of a chosen approach and persona pair (Fig.\ref{fig:algorithm_1_step_4}); we term the output {\em augmented few-shot examples}. This is abstracted in method \texttt{augment(.)}, where $\tilde{T}_{i,j}$ is the set of augmented few-shot examples corresponding to the approach and persona pair from ${s}_i$ and example $f_j$. An example is visualized in the bottom left of Fig.~\ref{fig:overall}, where the prompt contains different approaches for solving a math problem. 

\subsection{Designing the Prompts}
\label{subsec:techniques}

\noindent{\bf Step 5. Prompt Composition:} 
We create prompts for our approach using the best approach and persona pairs identified in step 3, and augmented few-shot examples from step 4 as shown in Fig.~\ref{fig:overall} and~\ref{fig:dev_prompt}.

We now describe two techniques to generate prompts with the augmented demonstrations (${T}$) that have been accumulated.

\vspace{1mm}
\noindent{\bf Candidate 1. \divse:} We first propose \divse (DIVerse reasoning path Self-Ensemble), a method to execute the diverse set of approaches in different inference calls and aggregate their solutions. Apart from the question to be solved and the augmented few-shot examples, the final prompt contains a persona, approach,  and additional instructions. One example is visualized in Fig.~\ref{fig:dev_prompt} (please refer to appendix for more examples of prompts: Fig.~\ref{fig:planning_baseline_prompt} through~\ref{fig:graph_coloring}). {\em Diversity is ensured through running inference with multiple prompts, each with a different approach and persona pairs and augmented few-shot examples.} However, since the approaches are executed separately, generating a solution (via aggregation of multiple responses) requires multiple inference calls, which can be costly. 

\vspace{1mm}
\noindent{\bf Candidate 2. \idivse:} To further reduce the inference costs while promoting diversity, we propose \idivse (In-call DIVerse reasoning path Self-Ensemble). In \idivse, {\em the final prompt is a composition of all approach and persona pairs and corresponding augmented few-shot examples, and the question to be solved}. An example is presented in Fig.~\ref{fig:overall} (bottom left). More examples of prompts are presented in the appendix in Fig.~\ref{fig:planning_baseline_prompt} through~\ref{fig:graph_coloring}. This noticeably decreases the number of calls to be made, since all few-shot examples are presented within the same prompt. We note that there might be error propagation due to the autoregressive nature of models. 
We evaluate this in detail in \S~\ref{error}.

\noindent{\bf Practicality.} Crucially, \auto finds approaches that are {\em general and reusable} across similar reasoning problems. We reused the strategies identified for solving \textsc{AQuA-RAT} and Planning benchmark respectively in the \textsc{MATH} (counting and probability) and Graph Coloring benchmarks. This also reduces the cost of repeated evaluation on a separate evaluation set.

\noindent{\bf Aggregation.} We aggregate the responses via majority vote for both prompting strategies. Other aggregation strategies can also be leveraged, such as utilizing the LLM itself to aggregate responses or weighted aggregation. In \S~\ref{ss:aggre}, we consider an aggregation strategy proposed by~\citet{yoran2023answering} and describe how compatible it is with our prompting approaches.

\section{Experiments}
\label{sec:experiments}
%\noindent{\bf Tasks \& Datasets.} 
We consider the following reasoning benchmarks.
%\begin{enumerate}[leftmargin=*,itemsep=0.0ex, parsep=0pt] 

\begin{figure*}[t]
    \centering
    \includegraphics[width=\textwidth]{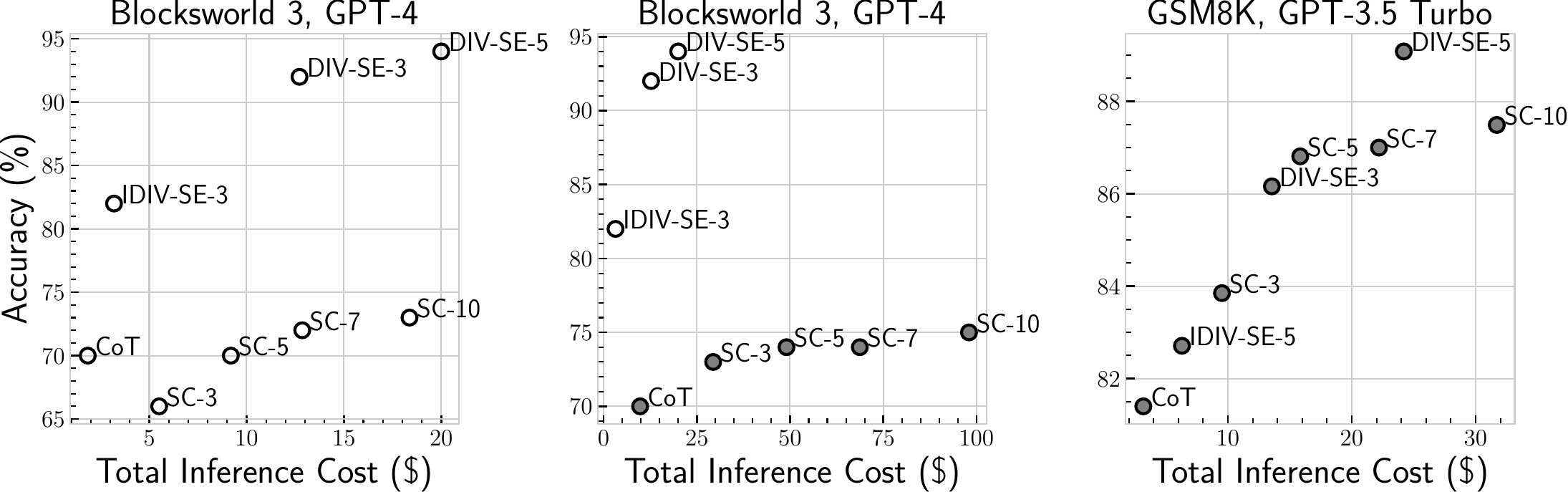}
    \caption{\textbf{Diversity of Thought enhances the inference cost and accuracy trade-off.} We compare \divse and \idivse with SC~\citep{wang2023selfconsistency} and CoT~\citep{wei2022chain} across three benchmarks. The x-axis indicates the total cost (as defined in \S~\ref{sec:experiments}) of running inference with the LLM on the benchmark using the given method, while the y-axis represents the LLM’s performance. The FS-CoT setting is represented by filled gray dots, while the ZS-CoT setting is indicated by unfilled dots. Notice that for \textsc{Blocksworld 3}, despite being in the ZS-CoT setting, our approaches are more performant than the SC-$s$ (FS-CoT) baseline.\vspace{-5mm}}
    \label{fig:cost_vs_accuracy_plot_mert2}
\end{figure*}

\noindent {\em Arithmetic Reasoning}:  We use: (i) \textsc{AQuA-RAT}~\citep{ling-etal-2017-program}, a suite of algebraic word problems,  (ii) \textsc{GSM8K}~\citep{cobbe2021training}, a benchmark of grade-school math problems described in natural language (involving elementary arithmetic operations), and (iii) \textsc{MATH} (Counting and Probability)~\citep{hendrycks2021measuring}, a collection of math problems from which we choose only counting and probability as these are not covered by \textsc{GSM8K} and \textsc{AQuA-RAT}. For all datasets, we use the \texttt{test} split for evaluation, containing 254, 1319, and 474 questions respectively. 

\noindent {\em Planning Capabilities}: We use the Blocksworld Planning benchmark proposed in \citet{valmeekam2022large, valmeekam2023planning}. 
The benchmark has two datasets: one involves 3 blocks (\textsc{Blocksworld 3}, 100 instances), while the other dataset involves 4 or 5 blocks (\textsc{Blocksworld 4/5}, 500 instances). 

\noindent {\em Constraint Satisfaction Optimization}: We use the \textsc{Graph coloring} benchmark~\citep{stechly2023gpt4} containing 100 examples to test reasoning for constraint satisfaction.
\noindent {\em Commonsense Reasoning}: We use \textsc{CommonsenseQA}~\citep{talmor-etal-2019-commonsenseqa} which consists of generic multiple-choice questions elicited for testing common sense reasoning. We use the \texttt{validation} split containing 1,221 questions.
%\end{enumerate}

\noindent{\bf Language Models.} We evaluate our proposed methods on both GPT-3.5 Turbo~\citep{chatgpt} and GPT-4~\citep{gpt4}. We also conduct an additional evaluation on LLaMA-2 70B~\citep{touvron2023llama2} to explore the performance of our technique on open-source LLMs. For the latter, we use \texttt{meta-llama/Llama-2-70b-chat-hf} through the Transformers library~\citep{wolf2019huggingface}.

\noindent{\bf Baselines.} We consider Chain-of-Thought (CoT)~\citep{wei2022chain} and Self-Consistency (SC~)~\citep{wang2023selfconsistency} as our baselines. For CoT, we consider two settings: zero-shot (ZS) CoT~\citep{kojima2022large} (i.e., ``\texttt{Think step by step}'' is added to the prompt), and few-shot (FS) CoT (i.e., CoT with demonstrations). In our SC runs, we set the temperature $T=0.7$ without top-$k$ truncation and sample up to $s \in [1,10]$ outputs (denoted SC-$s$). For all other approaches, we set $T=0$. We use ensembles of size 5 in \idivse and  \divse for \textsc{GSM8K} and \textsc{AQuA-RAT}. For the planning, \textsc{Graph Coloring}, and \textsc{CommonsenseQA} benchmarks, we use a size of 3. 

%\iffalse
\noindent{\bf Performance Metrics.} We measure the accuracy on the task, and the generation inference cost.To measure the cost, we assume 1000 tokens are about 750 words\footnote{https://openai.com/pricing}. For GPT-4 (8K) the input and output prices used to estimate inference cost are \$0.03/1k tokens and \$0.06/1k tokens, respectively. For GPT 3.5 Turbo (16K), the input and output prices used in the cost estimation are \$0.003/1k (tokens) and \$0.004/1k (tokens) respectively. 
%\fi

\noindent{\bf Results Summary.} include:
\noindent 
\noindent Across most benchmarks we consider, our techniques provide substantial performance gains (e.g., 16.52, 82.5, and 14.3 p.p improvements for \textsc{AQuA-RAT}, \textsc{Graph Coloring}, and \textsc{MATH} respectively). They are also Pareto optimal (in terms of the utility vs. cost trade-off). For the challenging planning benchmark (\textsc{Blocksworld 4/5}), our techniques improve accuracy by 29.6 p.p achieving {\em state-of-the-art} performance. Using GPT-4 for \textsc{Blocksworld 3}, our approach (in the ZS-CoT setting) is substantially more effective than SC-10 (in the FS-CoT setting) at 4$\times$ lower cost (Figure~\ref{fig:cost_vs_accuracy_plot_mert2} (center figure)). 

Since prompts are chained together in \idivse, error propagation is possible. Our evaluation on \textsc{AQuA-RAT} in ~\S~\ref{error} suggests that even though error propagation is estimated as less than 6.5\% for both models, these rates are comparable to differences in performance between \divse and \idivse. 
\noindent When combined with aggregation approaches that are capable of reasoning across the diverse generations~\citep{yoran2023answering}, we observe additional performance gains as shown in ~\S~\ref{ss:aggre}. For the \textsc{AQuA-RAT} benchmark for instance, we see an accuracy of 67.7\% for GPT-3.5 (3.23 p.p improvement to majority voting). 
%\end{enumerate}
%\vspace{-2mm}
\section{Results}
\label{sec:results}
\vspace{-2mm}
\subsection{Main Results}

We present the summary of results in Table~\ref{tab:results_1} and ~\ref{tab:results_2}. Detailed results are available in Appendix~\ref{appendix:addiitonal_results}. These also cover results on the impact of ensemble size in Appendix~\ref{appendix:ensemble_size}.

\begin{table}[t]
\centering
\small
\label{tab:comparison_cqa}
\begin{tabular}{cccc}
\toprule
  {\bf Method} & {\bf Graph Coloring} & {\bf BW 3} & {\bf BW 4/5 } \\
\midrule
 CoT  & 15.0 & 70.00   & 40.00\\
  SC-3 & 18.0 & 66.00   & 38.20\\
  SC-5 & 20.0 & 70.00   & 38.40 \\
  SC-7 & 22.0 &72.00   & 40.00 \\
  SC-10 & 23.0 & 73.00    &41.20 \\
      \idivse & 74.00 & 82.00   & 57.00 \\
      \divse & {\bf 97.00} & {\bf 94.00}  &  {\bf 69.60} \\

\bottomrule
\end{tabular}
\caption{Performance on \textsc{Graph Coloring} and  \textsc{Blocksworld} planning for GPT-4 in the ZS-CoT setting. We compare \divse and
\idivse with SC~\citep{wang2023selfconsistency} and CoT~\citep{wei2022chain}.}
\label{tab:results_1}
\end{table}

\begin{table}[ht]
\centering
\small
\label{tab:comparison_cqa}
\begin{tabular}{ccccccc}
\toprule
{\bf Setting} & {\bf Method} & {\bf AQuA} & {\bf MATH}  & {\bf CQA }\\
\midrule
\multirow{5}{*}{GPT-3.5 ZS} & CoT  & 59.00 & 31.90   &  71.40\\
 & SC-3 & 61.40  & 32.07   &  72.00\\
   & SC-5 & 63.37 & 38.19  &   72.80\\
     & \idivse & 62.60 & 42.50   &  74.00\\
     & \divse & {\bf 72.83} & {\bf 44.94}   & {\bf 74.50} \\ \hline
% & &  &   &  & \\
     \multirow{3}{*}{GPT-3.5 FS} & CoT  & 57.48 & 30.38   &  79.4\\
     & \idivse & 64.57 & 44.10 &  80.00  \\
     & \divse & {\bf 72.84} & {\bf 52.22} &  {\bf 80.40}  \\
\midrule
    \multirow{3}{*}{GPT-4 ZS} &  CoT  & 70.47 & 62.24    & 81.60\\
     & \idivse & 71.65 & 72.00   &  {\bf 82.50 }\\
     & \divse & {\bf 80.31} & {\bf 79.11}    & 81.70\\ \hline
    % & &  &   & & \\
  \multirow{3}{*}{GPT-4 FS}  & CoT  & 71.90 & 66.46  & 87.70\\
     & \idivse & 79.90 & 72.00 &  {\bf 89.00 } \\
     & \divse & {\bf 84.25} & {\bf 80.76}   & 88.00\\
\bottomrule
\end{tabular}
\caption{Performance on \textsc{AQuA-RAT}, \textsc{MATH} (Counting and Probability),  and \textsc{CommonsenseQA} for GPT-3.5 Turbo and GPT-4 in the ZS-CoT and few-shot-CoT settings respectively. \vspace{-5mm} 
}
\label{tab:results_2}
\end{table}

\subsubsection{Arithmetic reasoning via \textsc{AQuA-RAT}} 
\noindent{\em GPT-4 Results:} In Table~\ref{tab:results_2}, we observe that \divse achieves an accuracy increase of 9.84 and 14.6 p.p in the FS-CoT (baseline accuracy of 71.9\%) and ZS-CoT (baseline of 70.47\%) settings, respectively. While the gains from \idivse are nominal in ZS-CoT, it achieves a boost of 7.7 p.p for FS-CoT.

\noindent{\em GPT-3.5 Results:} In Table~\ref{tab:results_2}, we see that \divse yields a gain of 14.23 and 16.52 p.p in the FS-CoT (baseline of 57.48\%) and ZS-CoT (baseline of 59\%) settings, respectively. Within the FS-CoT setting, \idivse gets an absolute increase of 7 p.p.

Note that Fig.~\ref{fig:cost_vs_accuracy_plot_mert} also displays the total inference cost. Both \idivse and \divse are {\em Pareto optimal}, indicating their capacity to achieve a higher accuracy while maintaining low costs.

\subsubsection{Counting and probabilistic reasoning via \textsc{MATH}}

\noindent{\em GPT-4 Results:} From Table~\ref{tab:results_2}, we see that \divse achieves an accuracy increase of 14.3 and 16.87 p.p in the FS-CoT (baseline of  66.46\%) and ZS-CoT (baseline of 62.24\%) settings, respectively. On the other hand, \idivse achieves a boost of 5.54 and 9.76 p.p in the FS-CoT and ZS-CoT settings, respectively, over the baseline.

\noindent{\em GPT-3.5 Results:} From Table~\ref{tab:results_2}, we see that \divse yields a gain of 21.84 and 13.04 p.p in the FS-CoT (baseline of 30.38\%) and ZS-CoT (baseline of 31.90\%) settings, respectively. Likewise \idivse achieves a boost of 13.72 and 10.60 p.p in the FS-CoT and ZS-CoT settings, respectively.

\subsubsection{Planning via \textsc{Blocksworld}}

 \noindent{\em Setup:} The benchmark provides both natural language and Planning Definition and Domain Language prompts~\citep{mcdermott1998pddl}.
 We use natural language prompts in all the experiments. For the baseline runs, we introduce minor alterations to the prompt originally proposed by~\citet{valmeekam2023planning}. These changes involve incorporating an explicit directive to prevent under-block movement and resolving minor language ambiguities we observed to be problematic during initial investigation. Furthermore, we reposition the initial condition and goal state information to the beginning of the prompt. The modified improved prompt is presented in Fig.~\ref{fig:planning_baseline_prompt}. 
 
We aggregate the plans through majority voting and utilize string matching for comparing the plans. As a result, we optimize the plan by eliminating the redundant ``\texttt{no-op}'' steps.

\begin{comment}
\begin{Verbatim}[commandchars=\\\{\}]
\centering
[(unstack b c), (stack a b), (unstack a b), (stack b a)] ==
[(unstack b c), \sout{(stack a b)}, \sout{(unstack a b)}, (stack b a)] 
\end{Verbatim}
\end{comment}

\noindent{\em GPT-4 Results:} We note that GPT-4 performs slightly better in a ZS setting, and use this to run all experiments. From Fig.~\ref{fig:cost_vs_accuracy_plot_mert}, we observe that for \textsc{Blocksworld 3}, ZS-CoT records an accuracy of 70\%, while SC-10 reaches an accuracy level of 73\%. \idivse enhances the absolute accuracy by 12 p.p above the ZS-CoT baseline, while \divse produces an impressive {\em state-of-the-art} accuracy of 94\%. An analysis of the six unsuccessful instances suggests the capacity for further performance improvement by increasing the size of the ensemble, as already two out of five current approaches generate accurate plans. For the \textsc{Blocksworld 4/5} case, the ZS-CoT accuracy is 40\%, while SC-10 has an accuracy of 41.2\%. Here, \idivse results in an absolute gain of 17 p.p above the ZS-CoT baseline, and \divse too enhances performance, leading to 69.6\%. As outlined in Fig.~\ref{fig:cost_vs_accuracy_plot_mert} and ~\ref{fig:cost_vs_accuracy_plot_mert2}, both \idivse and \divse achieve {\em Pareto optimality}. 

\noindent{\em GPT-3.5 Results:} The baseline performance on \textsc{Blocksworld 3} is 6\%, and on \textsc{Blocksworld 4/5} is 0.6\%. We do not see any additional improvement using both \idivse and \divse. Qualitatively, we observe that during plan generation, GPT-3.5 fails to follow the restrictions provided as part of the problem instructions too often, leading to either infeasible or incorrect plans. This shows instruction following capabilities are crucial to the success of the methods proposed here.

\subsubsection{Constraint Satisfaction via \textsc{graph Coloring}}
There may exist numerous non-optimal yet valid colorings for a given graph. Since exact string matching is not usable for identifying the majority solution from the ensembles of \idivse and \divse, we employ the external, sound verifier~\citep{stechly2023gpt4} to pick the correct solution.

\noindent{\em GPT-4 Results:} From Fig.~\ref{fig:cost_vs_accuracy_plot_mert}, it is observed that ZS-CoT achieves an accuracy of 15\%, whereas SC-10 attains an accuracy level of 23\%. \idivse improves the absolute accuracy by 59 p.p above the ZS-CoT baseline. Remarkably, \divse delivers a {\em state-of-the-art} accuracy of 97\%. Given that GPT-4's performance plateaus in the ZS setting, we  chose to omit conducting the few-shot experiments.

\noindent{\bf Summary:} Methods in this work often demonstrate state-of-the-art performance on reasoning tasks. This is most significant in the planning and constraint satisfaction benchmarks, where the corresponding authors claimed immense difficulty for existing LLMs. Our work shows that status-quo prompt design approaches including chain of thought are too generic for these problems, and prompt customization (via \auto) can yield substantial gains by guiding the chain of thought to the general nature of the problem.
%\end{minipage}}

\subsection{Open Source Models}

Due to the limited computational budget, we only performed experiments with the \textsc{AQuA-RAT} benchmark. Please refer to Appendix~\ref{appendix:open-source} for further details. Table~\ref{tab:open-source} demonstrates the results for LLaMA-2 70B with 8-bit quantization. \divse and \idivse demonstrate an improvement of over 10 p.p over the baseline in the FS-CoT settings. However, the gain in the ZS-CoT setting has been negligible. We hypothesize that this is partly due to model's lack of capabilities to both follow instructions and the mentioned approach in the absence of examples. 
\begin{table}[!ht]
  \centering
    \small
  \label{tab:comparison_prompting_strategy}
  \begin{tabular}{lcc}
    \toprule
     {\bf Prompting Strategy} & {\bf ZS-CoT (\%)} & {\bf FS-CoT (\%)} \\
    \midrule
    CoT & 31.32 & 29.1 \\
    \idivse & 27.00 & 39.7 \\
    \divse & \bf{32.00} & \bf{39.9} \\

    \bottomrule
  \end{tabular}
      \caption{Results on \textsc{AQuA-RAT} and LLaMA-2 70B.\vspace{-5mm}}
      \label{tab:open-source}
\end{table}

\subsection{Errors \& Prompt Utility} 
\label{error}

\noindent{\em Error Propagation:} Due to the autoregressive nature of LLM decoding, early incorrect answers in \idivse may get propagated to the latter ones. To quantify this, we select examples where the solution is incorrect and all five approaches produce the same erroneous answer. We focus only on these cases to see if e.g., a wrong conclusion in the initial approaches leaks into the following ones. Next, we attempt the last two approaches again in a separate session: if the LLM generates the same outcomes as in the original session (i.e., \idivse setup) within 3 attempts, we consider it as no error propagation. However, if it does not produce the same answer within the 3 attempts, we interpret this as a case of error propagation since the change in answer could be attributed to the initial approaches with wrong answers in the chain. We measure this phenomenon on \textsc{AQuA-RAT} (FS-CoT) on both GPT-4 and GPT-3.5. We find that GPT-4 and GPT-3.5 have error propagation rates of 6.2\% and 5.5\% respectively, which are comparable to performance differences between \divse and \idivse, {\em making error propagation one of the main explanatory hypotheses for the differences between the two methods}. Reducing these error rates remains a challenging problem given the autoregressive nature of current LLMs.

\begin{table*}[!ht]
  \centering
\small
  \begin{tabular}{cll}
\toprule
    {\bf Dataset, Model} & {\bf Persona, Approach} & {\bf Accuracy (\%)}  \\
\midrule
    \multirow{3}{*}{\textsc{AQuA-RAT}, GPT-3.5} & $\emptyset$,Think step by step & 57.48  \\
    & $\emptyset$, Using Algebra & 60.24 \footnotesize{\bf (+2.76)} \\
    & Thinking like Alan Turing, $\emptyset$ & 61.81 \footnotesize{\bf (+4.33)}  \\
        & Dr. Patel: A renowned mathematician, $\emptyset$ & {\bf 65.75 \footnotesize{(+8.27)}}  \\
    \hline
    \multirow{3}{*}{\textsc{Blocksworld 4/5}, GPT-4} & $\emptyset$, State tracking prompt~\citep{valmeekam2022large} & 42.00  \\
    & \textcolor{blue}{$\emptyset$, Finite State Machine} & 55.80 \footnotesize{\bf (+13.80)} \\
    & \textcolor{blue}{Alan Turing, Action Rationale} & 57.80 \footnotesize{\bf (+15.80)}  \\
  & \textcolor{blue}{Alan Turing, Progressive Block Placement Approach} & {\bf 58.80 \footnotesize{(+16.80)}}  \\
\bottomrule
  \end{tabular}
    \caption{\textbf{Prompts, derived from approaches and personas, boost performance.} \textcolor{blue}{Blue} rows denote ZS-CoT prompts, while black lines denote FS-CoT prompts. $\emptyset$ denotes absence (of persona or approach respectively). 
    }
 \label{fig:personas_approaches}
\end{table*}

\noindent{\em Beyond Thinking Step by Step:} The diverse approaches and personas we utilize not only enhance the performance in \idivse and \idivse, but are also independently superior to ZS-CoT. Table~\ref{fig:personas_approaches} highlights this effect, which showcases the importance of conditioning the model for solutions via \auto.

\subsection{Alternative Aggregation Strategies}
\label{ss:aggre}

\begin{table}[t]
  \centering
    \small
  \label{tab:comparison_agg_strategy}
  \begin{tabular}{lcc}
    \toprule
     {\bf Method} & {\bf GPT-4 (\%)} & {\bf GPT-3.5 (\%)} \\
    \midrule
    Majority Voting & {\bf 79.90} & 64.47 \\
    Meta Reasoning & {79.24} & {\bf 67.70} \\
    \bottomrule
  \end{tabular}
      \caption{{\bf Alternative aggregation strategies.} Observe that, for the \textsc{AQuA-RAT} benchmark (FS-CoT), \idivse produces more accurate results only with GPT-3.5. \vspace{-5mm}}
      \label{tab:meta}
\end{table}

Our aggregation thus far relies on majority voting. Alternatively, we can also utilize the meta reasoning technique proposed by~\citet{yoran2023answering} to accumulate the results and exploit the rich information present in the reasoning steps.
To this end, we store the responses generated by \idivse, and request the model to meta reason over them in a different prompt and session. Table~\ref{tab:meta} suggests that the proposed reasoning paths contain rich information that is effectively exploited by the meta reasoning aggregation. Future post-hoc  techniques may consider to learn the accuracy of the diverse prompting approaches, and weigh them accordingly. Nevertheless, the fact that techniques presented here provide visible improvements even with simple approaches like majority voting, demonstrates their added value independently from different aggregation algorithms.

\section{Related Work}
\label{sec:related}
\noindent{\bf Prompt Optimization:} \citet{pryzant2023automatic} models the prompts as optimizable discrete variables, and minimizes the loss of the reasoning task. \citet{jones2023automatically} optimize over the prompt space, but to identify failure modes. However, optimization-based approaches often require the task to have a differentiable loss function, which is a strong condition. In our work, we utilize feedback from the LLM (not through gradients) during prompt design. Similarly to~\citet{cheng2023batch}, \idivse batches the responses for multiple queries within a prompt.

\noindent{\bf Decoding Optimizations and Tools:}~\citet{wang2023selfconsistency} replace the naive greedy decoding by sampling a diverse set of reasoning paths (e.g., through temperature sampling), and then selects the most consistent answer. ~\citet{chen2022program} express the reasoning process as a program, which is then delegated to an external tool. 
In our work, we neither change the decoding process nor assume the existence of trusted tools. This makes our solution directly applicable to black-box models.

\noindent{\bf Prompting Strategies:} ~\citet{brown2020language} note that demonstrations to prompts, encoded as input-output pairs, produce drastic performance increase in larger LLMs.~\citet{wei2022chain} encourage internal dialogue by forcing the LLM to generate a sequence of intermediate steps for reasoning problems. This improves reasoning performance on larger LLMs~\citep{nye2021show,chung2022scaling, kojima2022large}. ~\citet{zhou2022least} automatically break a complex problem into simpler sub-problems and then solve them in sequence. Across all these techniques, the common practice is to keep the prompts fixed, but aggregate responses across multiple trials by varying the temperature. In our work, we vary the input prompt itself. A work that is similar in spirit is that of~\citet{yoran2023answering}, which instead of aggregating the response of multiple reasoning paths, forces the model to reason across them before aggregation. Another relevant work is that of \citet{li2023making}, which shows the importance of prompt diversity. However, they rely on selecting few-shot demonstrations from a hold-out set (which defines diversity in their method), without explicitly stating reasoning pathways.

\section{Conclusions}
In this work, we promoted diversity of thought as a principled prompting strategy and proposed methodologies that leverage the LLM as a guide to design a diverse set of approaches to solve complex reasoning tasks. Extracting solution approaches from LLMs themselves becomes a discovery mechanism that seeds and conditions generative solutions. 
Reported results on a variety of tasks confirm that there is a large space for improvement in complex reasoning by uncovering the necessary skills and knowledge from LLMs through targeted and diverse prompting methods. These results demonstrated how promoting diversity can improve the Pareto frontier of accuracy-cost trade-off for current LLMs and yield {\em state-of-the-art} solutions for planning and mathematical reasoning tasks. We hope that future work will expand these results to complex tasks from other real-world applications.

\section{Limitations}
Our study mainly experimented with GPT-3.5 and GPT-4 models because of their instruction-following capabilities. While current open-source models have shown remarkable improvements to this end, they are still not able to reliably follow instructions relevant to complex reasoning tasks (e.g. state tracking, plan validity, constraint satisfaction). We hope that progress in the field will enable further experimentation in this direction. 

In addition, we also observe that error propagation during autoregressive generation may sometimes negatively impact the performance of \idivse, where all approaches are executed in order within the same prompt. Some of this could be addressed by explicitly instructing the model to forget about the previous solution but ultimately as long as previous generation history remains in context and short-term memory, error propagation risks may still need to be tracked and measured.

\bibliography{custom}

\newpage

\appendix
\section*{Appendix}

\iffalse
\newtcolorbox{userinput}{
    colback=gray!10,
    colframe=gray!40,
    fonttitle=\bfseries,
    coltitle=black,
    colbacktitle=gray!60,
    enhanced,
    drop shadow=black!5!white,
    left=8mm,
    right=8mm,
    top=3mm,
    bottom=3mm,
    boxsep=0mm,
    sharp corners=south,
    rounded corners=north,
    title=Prompt:
    
    }
\fi

\section{Prompt used for \auto}
\begin{figure*}[h]
    \centering
    \includegraphics[width=0.9\textwidth]{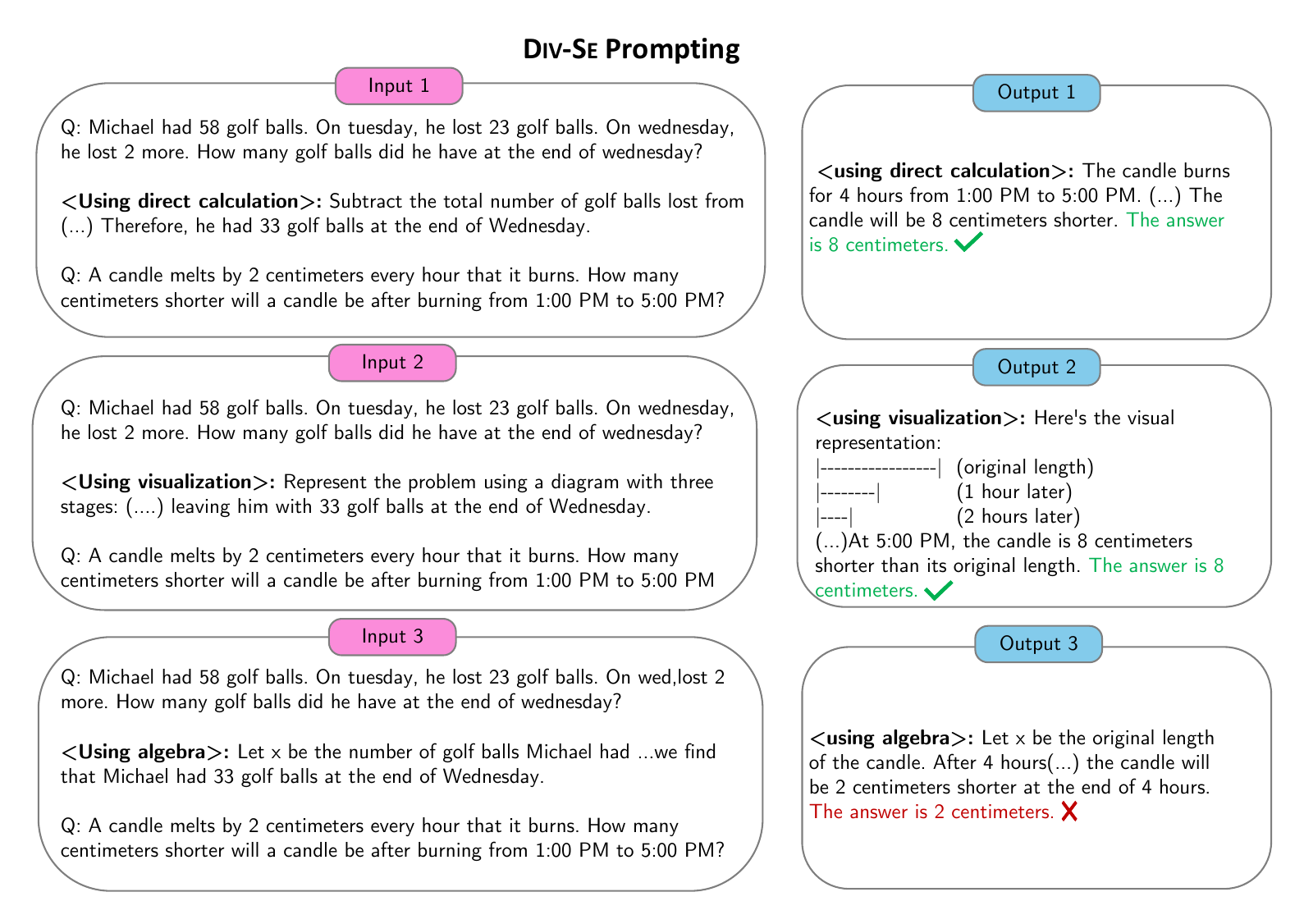}
    \caption{ \divse prompting.}
    \label{fig:dev_prompt}
\end{figure*}

Our diverse prompting strategy for \idivse and \divse is showcased in Fig.~\ref{fig:overall} and Fig.~\ref{fig:dev_prompt} respectively. The instrumental prompt template that determines our approaches is presented in Fig.~\ref{fig:auto-example}.

\section{Model Details}
\label{appendix:open-source}

\subsection{Open-Source Models}
We perform the Llama-2 70B experiments with a single 80GB A100 GPU. To fit the 70B model to a single A100, we use 8-bit precision through \texttt{bitsandbytes}~\citep{dettmers2022llm, dettmers2022optimizers}. Further, \citet{dettmers2022llm} reports no performance drop with this quantization method.

As the system prompt, we use \texttt{You are a helpful, respectful and honest assistant.} We perform inference with greedy decoding, having temperature $T = 0$. 

\section{ Additional Results}
\label{appendix:addiitonal_results}
In this section, we provide additional results on \textsc{CommonsenseQA} and \textsc{GSM8K} benchmarks.

\subsection{Common sense via \textsc{CommonsenseQA}}

Table~\ref{tab:results_2} presents the results of the experiments. Overall, the improvements in accuracy are relatively modest. This is likely because answering questions in \textsc{CommonsenseQA} does not demand as much reasoning and thought diversity as is required in some other benchmarks. In addition, the dataset also contains a number of ambiguous questions, which if read verbatim may have many plausible answers but the ground truth contains only one answer.

\begin{comment}

\end{comment}

\subsection{Arithmetic reseasoning via \textsc{GSM8K}} 

\noindent\underline{\em GPT-4 Results:} As shown in Fig.~\ref{fig:gsm_turbo_gpt_4}, accuracy on \textsc{GSM8K} have nearly plateaued, with the ZS-CoT and FS-CoT baselines achieving accuracies of 94\% and 95\% respectively. \idivse does not produce any significant gains in either setting. On the other hand, \divse reaches accuracy of 96.3\% in both FS-CoT and ZS-CoT settings, providing a modest improvement. 

\noindent\underline{\em GPT-3.5 Results:} Here, the gains are more substantial. Compared to the ZS-CoT baseline of 76.11\%, \idivse provides an improvement of 5.31 p.p. \divse goes a step further, enhancing the accuracy by 10.39 p.p. In the FS-CoT setting, \divse posts an accuracy improvement of 7.68 p.p (with a baseline accuracy of 81.4\%).

Fig.~\ref{fig:cost_vs_accuracy_plot_mert2} (rightmost) presents the cost vs. accuracy trade-offs between \idivse, \divse, and SC. While the performance of SC does improve with the expansion of reasoning paths, both \idivse and \divse offer better trade-offs. 

\section{Evaluating Ensemble Sizes}
\label{appendix:ensemble_size}
\begin{figure*}[t]
    \centering
    \includegraphics[width=\textwidth]{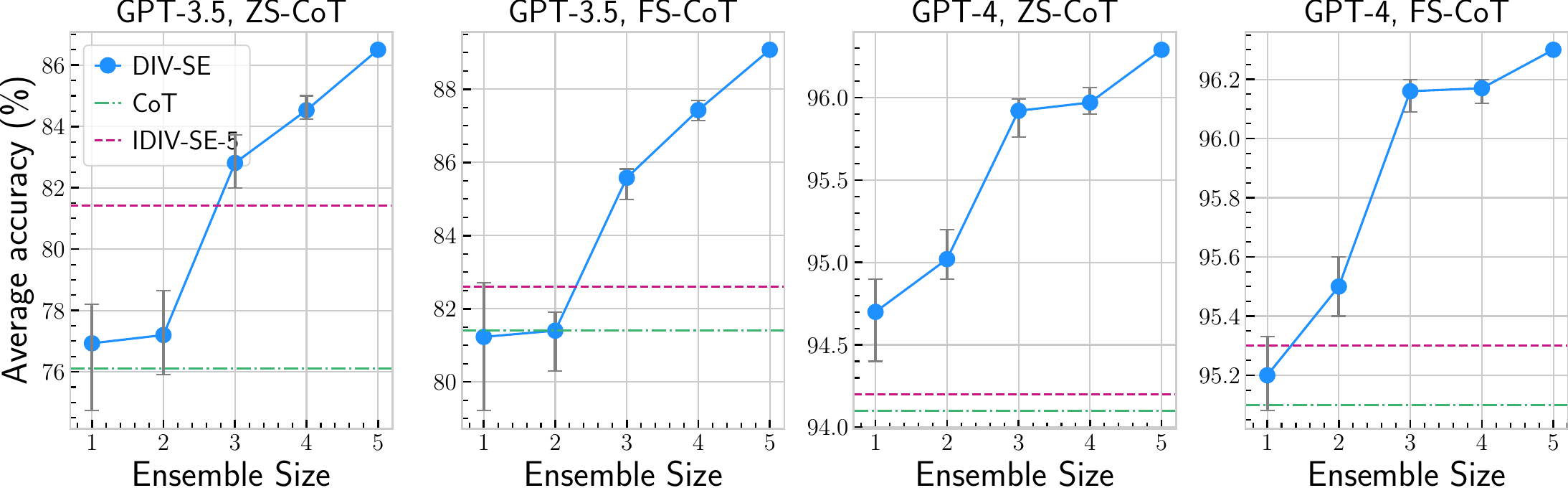}
    \caption{{\bf Average accuracy} for different ensemble sizes on \textsc{GSM8K} for ZS-CoT and FS-CoT settings on GPT-4 and GPT-3.5. Note that all graphs are zoomed in. }
    \label{fig:gsm_turbo_gpt_4}
\end{figure*}

\begin{figure*}[t]
    \centering
    \includegraphics[width=\textwidth]{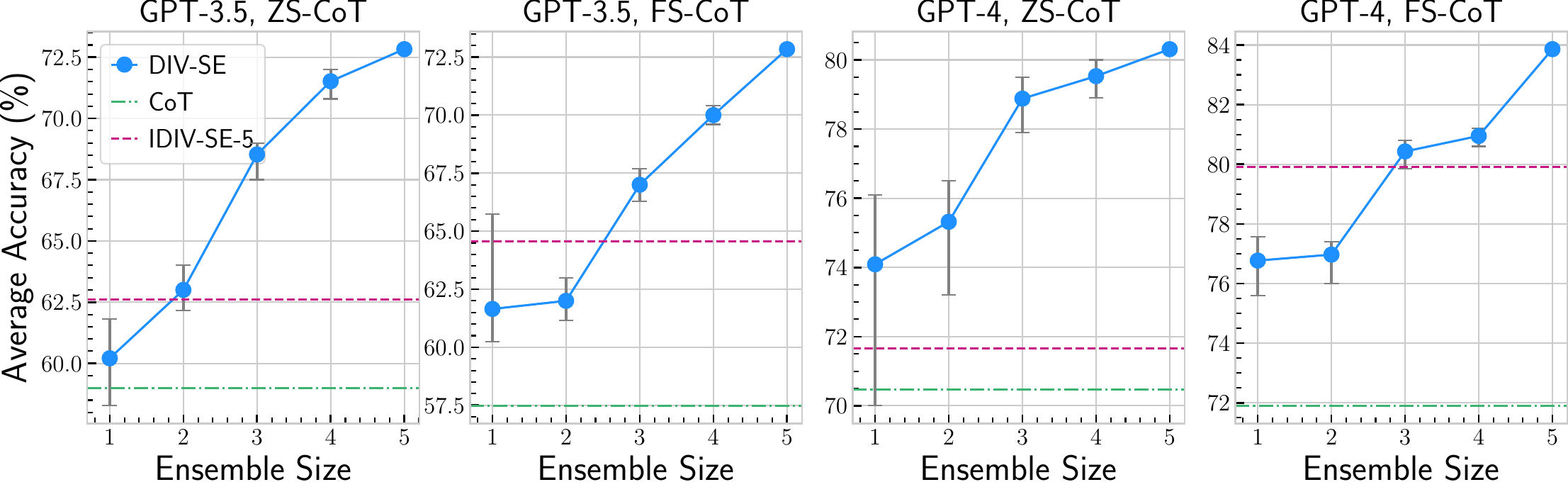}
    \caption{{\bf Average accuracy} for different ensemble sizes on \textsc{AQUA-RAT} for ZS-CoT and FS-CoT settings on GPT-4 and GPT-3.5. Note that all graphs are zoomed in.}
    \label{fig:acqua_turbo_gpt_4}
\end{figure*}

Figure~\ref{fig:gsm_turbo_gpt_4} depicts the average accuracy of different ensemble sizes on GSM8K for both ZS-CoT and FS-CoT settings, utilizing GPT-4 and GPT-3.5.
Similarly, Figure~\ref{fig:acqua_turbo_gpt_4} demonstrates the average accuracy of various ensemble sizes on AquA for both ZS-CoT and FS-CoT settings, using GPT-4 and GPT-3.5. 
It is noteworthy that in both AQuA and GSM8K, even an ensemble of size three yields significant performance improvements over the baseline, which we attribute to the high diversity and independence of reasoning paths.

\section{Prompt Templates}
The following section provides a comprehensive visual representation of the prompts used in our study. These prompts, depicted in Figures~\ref{fig:planning_baseline_prompt} through ~\ref{fig:graph_coloring}, were used in different settings and for the planning, AQuA, and graph coloring benchmarks, and incorporate various personas and approaches.

Figure~\ref{fig:planning_baseline_prompt} illustrates the prompt used in the baseline run, Figure~\ref{fig:planning_baseline_prompt_finite_state_machine} demonstrates the prompt employed when applying a Finite State Machine approach, Figure~\ref{fig:planning_baseline_prompt_alan_turing_rationale} depicts the prompt used when incorporating the persona of Alan Turing and an Action Rationale approach, and lastly, Figure~\ref{fig:planning_baseline_prompt_alan_turing_progressive} shows the prompt used when applying the persona of Alan Turing and the Progressive Block Placement approach in the zero-shot setting for the planning benchmark.

Figure~\ref{fig:aqua_few_shot_prompt_algebra_approach} illustrates the algebraic approach, while Figure~\ref{fig:aqua_few_shot_prompt_algebra_approach_alan_turing} and Figure~\ref{fig:aqua_few_shot_prompt_algebra_approach_dr_patel} demonstrate the prompts incorporating the personas of Alan Turing and Dr. Patel, a renowned mathematician, respectively, in the few-shot-CoT setting for the AQuA benchmark.

In Figure~\ref{fig:graph_coloring} we present the \idivse prompt used in the zero-shot setting for the graph coloring benchmark.

\definecolor{lightgray}{RGB}{240,240,240} %
\renewcommand{\sfdefault}{cmss} % Computer Modern Sans Serif
\begin{figure*}[!th]
\begin{tcolorbox}[
    enhanced,
    opacityback=0.2,
    opacityframe=0.2,
    colback=lightgray,
    colframe=lightgray, 
    top=1mm, 
    bottom=1mm, 
    left=1mm, 
    right=1mm,
    %sharp corners,
    fontupper=\sffamily\small,
    ]    
\begin{small}
\noindent

Use five distinct approaches to solve the given problem accurately. If there is no exact match choose the closest option.

Q: \{Question\}

Use the following output format:

Approach 1 {< name of the approach >} : {< Details of Approach 1} >

Approach 2 {< name of the approach >} : {< Details of Approach 2} >

Approach 3 {< name of the approach >} : {< Details of Approach 3} >

Approach 4 {< name of the approach >} : {< Details of Approach 4} >

Approach 5 {< name of the approach >} : {< Details of Approach 5} >
\noindent
\end{small}
\end{tcolorbox}
\caption{Prompt template for extracting diverse approaches for problem solving.}
\label{fig:auto-example}
\end{figure*}

\clearpage

\begin{comment}

\end{comment}

\begin{comment}

\end{comment}

\begin{comment}

\end{comment}

\begin{comment}

\end{comment}

\begin{comment}

\end{comment}

\begin{comment}

\end{comment}

\renewcommand{\sfdefault}{cmss}
\begin{figure*}[!th]
\begin{tcolorbox}[
    enhanced,
    opacityback=0.2,
    opacityframe=0.2,
    colback=lightgray,
    colframe=lightgray, 
    top=1mm, 
    bottom=1mm, 
    left=1mm, 
    right=1mm,
    %sharp corners,
    fontupper=\sffamily\small,
    ]    
\begin{large}
\noindent
\textbf{\textsc{Input Prompt}} \newline \newline
Solve the problem accurately using \{approach name\}.  \newline \newline
Question: John found that the average of 15 numbers is 40. If 10 is added to each number then the mean of the numbers \newline
Answer Choices: (A) 50 (B) 45 (C) 65 (D) 78 (E) 64 \newline
A: If 10 is added to each number, then the mean of the numbers also increases by 10. So the new mean would be 50. The answer is (A). \newline \newline
Approach <using \{approach name\} >:  \newline \newline

\textbf{\textsc{Augmented Few-shot Examples}}\newline \newline
\textbf{Approach <using visualization>}: \newline
To solve this problem using visualization, let's consider the given information and the effect of adding 10 to each number on the average. \newline
1.	The average of 15 numbers is 40. This means that the sum of these 15 numbers is 15 * 40 = 600. \newline
2.	Now, we add 10 to each number. Since there are 15 numbers, the total added is 15 * 10 = 150. \newline
3.	The new sum of the 15 numbers is 600 + 150 = 750. \newline
4.	To find the new average, we divide the new sum by the number of values, which is 750 / 15 = 50. 
So, the mean of the numbers after adding 10 to each number is 50. The answer is: (A) 50 \newline 

\textbf{Approach <using logical reasoning>}:  \newline
We know that adding the same constant to each number in a set will increase the average by that constant. In this case, we are adding 10 to each number, so the average will increase by 10.
New average = 40 + 10 = 50
The answer is: (A) 50 \newline

\textbf{Approach <using elimination strategy>}:  \newline
We know that adding 10 to each number will increase the average by 10. So, we're looking for an answer choice that is 10 more than the original average of 40.\newline
(A) 50 is 10 more than 40. \newline
(B) 45 is only 5 more than 40. \newline
(C) 65 is 25 more than 40. \newline
(D) 78 is 38 more than 40.\newline 
(E) 64 is 24 more than 40.\newline
Only (A) 50 is exactly 10 more than the original average of 40, so that's the correct answer. The other choices can be eliminated because they are not 10 more than the original average.\newline

\textbf{Approach <using algebra>}:  \newline
Let's denote the sum of the 15 numbers as S. Since the average of the 15 numbers is 40, we can write the equation:
S / 15 = 40
S = 40 * 15 = 600
Now, we add 10 to each number, so the new sum is S + 15 * 10 = 600 + 150 = 750. The new average is:
New average = 750 / 15 = 50
The answer is: (A) 50

\noindent
\end{large}
\end{tcolorbox}
\caption{Illustration of Augmentation of few-shot examples -  Algorithm~\ref{alg_1}, Step 4}
\label{fig:algorithm_1_step_4}
\end{figure*}

\renewcommand{\sfdefault}{cmss}
\begin{figure*}[!th]
\begin{tcolorbox}[
    enhanced,
    opacityback=0.2,
    opacityframe=0.2,
    colback=lightgray,
    colframe=lightgray, 
    top=1mm, 
    bottom=1mm, 
    left=1mm, 
    right=1mm,
    %sharp corners,
    fontupper=\sffamily\small,
    ]    
\begin{large}
\noindent
I am playing with a set of blocks where I need to arrange the blocks into stacks.\newline
\newline
[STATEMENT]\newline
As initial conditions I have that, the orange block is clear, the hand is empty, the blue block is on top of the red block, the orange block is on top of the blue block and the red block is on the table.
My goal is to have that the red block on top of the blue block and the orange block on top of the red block.
\newline\newline
Here are the actions I can do:\newline
Pick up a block from the table\newline
Unstack a block from on top of another block\newline 
Put down a block on the table\newline
Stack a block on top of another block\newline
\newline
I have the following restrictions on my actions:\newline
I can only pick up or unstack one block at a time.\newline
I can only pick up or unstack a block if my hand is empty.\newline
I can only pick up a block if the block is on the table and the block is clear. A block is clear if the block has no other blocks on top of it and if the block is not picked up.\newline
I can only unstack a block from on top of another block if the block I am unstacking was really on top of the other block.\newline
I can only unstack a block from on top of another block if the block I am unstacking is clear.\newline
Once I pick up or unstack a block, I am holding the block.\newline
I can only put down a block that I am holding.\newline
I can only stack a block on top and not under of another block if I am holding the block being stacked.\newline
I can only stack a block on top and not under of another block if the block onto which I am stacking the block is clear.\newline
Once I put down or stack a block, my hand becomes empty.\newline
Once you stack a block on top of a second block, the second block is no longer clear.\newline
\newline
What is the plan to achieve my goal? Just give the actions in the plan.
\newline\newline
[PLAN]

\noindent
\end{large}
\end{tcolorbox}
\caption{Zero-shot prompt used in the baseline run of the Planning - Blocksworld Domain}
\label{fig:planning_baseline_prompt}
\end{figure*}

\begin{figure*}[!th]
\begin{tcolorbox}[
    enhanced,
    opacityback=0.2,
    opacityframe=0.2,
    colback=lightgray,
    colframe=lightgray, 
    top=1mm, 
    bottom=1mm, 
    left=1mm, 
    right=1mm,
    %sharp corners,
    fontupper=\sffamily\small,
    ]    
\begin{large}
    
\noindent
You are playing with a set of blocks where you need to arrange the blocks into stacks. What is the plan to achieve the goal?\newline\newline
<Initial State> : As initial conditions you have that, the orange block is clear, the hand is empty, the blue block is on top of the red block, the orange block is on top of the blue block and the red block is on the table.\newline
<Goal State> : Your goal is to have that the red block on top of the blue block and the orange block on top of the red block.\newline\newline
Here are the actions you can do:\newline
-Pick up a block from the table\newline
-Unstack a block from on top of another block\newline
-Put down a block on the table\newline
-Stack a block on top of another block\newline\newline
Rules:\newline
1. You can only pick up or unstack one block at a time.\newline
2. You can only pick up or unstack a block if your hand is empty.\newline
3. You can only pick up a block if the block is on the table and the block is clear. A block is clear if the block has no other blocks on top of it and if the block is not picked up.\newline
4. You can only unstack a block from on top of another block if the block you are unstacking was really on top of the other block.\newline
5. You can only unstack a block from on top of another block if the block you are unstacking is clear.\newline
6. Once you pick up or unstack a block, you are holding the block.\newline
7. You can only put down a block that you are holding.\newline
8. You can only stack a block on top and not under of another block if you are holding the block being stacked.\newline
9. You can only stack a block on top and not under of another block if the block onto which you are stacking the block is clear. \newline  
10. Once you put down or stack a block, your hand becomes empty.\newline
11. Once you stack a block on top of a second block, the second block is no longer clear.\newline\newline
Using a finite state machine and a search algorithm what is the plan to achieve the goal? You can model each state of the blocks configuration on the table and the hand as a state. For each action step check that the step follows the rules and that the step brings you closer to the goal. After each action describe the state of the table and hand. Always check whether the final state satisfies the goal mentioned. <Goal State> : Your goal is to have that the red block on top of the blue block and the orange block on top of the red block.\newline\newline
[PLAN]
\noindent
\end{large}
\end{tcolorbox}
\caption{The Zero-shot prompt using Finite State Machine Approach for solving the Planning - Blocksworld Domain Problem.}
\label{fig:planning_baseline_prompt_finite_state_machine}
\end{figure*}

\begin{figure*}[!th]
\begin{tcolorbox}[
    enhanced,
    opacityback=0.2,
    opacityframe=0.2,
    colback=lightgray,
    colframe=lightgray, 
    top=1mm, 
    bottom=1mm, 
    left=1mm, 
    right=1mm,
    %sharp corners,
    fontupper=\sffamily\small,
    ]    
\begin{large}
\noindent
You are playing with a set of blocks where you need to arrange the blocks into stacks.\newline\newline
<Initial State> : As initial conditions you have that, the orange block is clear, the hand is empty, the blue block is on top of the red block, the orange block is on top of the blue block and the red block is on the table.\newline\newline
<Goal State> : Your goal is to have that the red block on top of the blue block and the orange block on top of the red block.\newline\newline
Here are the actions you can do:\newline
-Pick up a block from the table\newline
-Unstack a block from on top of another block\newline
-Put down a block on the table\newline
-Stack a block on top of another block\newline\newline
Rules:\newline
1. You can only pick up or unstack one block at a time.\newline
2. You can only pick up or unstack a block if your hand is empty.\newline
3. You can only pick up a block if the block is on the table and the block is clear. A block is clear if the block has no other blocks on top of it and if the block is not picked up.\newline
4. You can only unstack a block from on top of another block if the block you are unstacking was really on top of the other block.\newline
5. You can only unstack a block from on top of another block if the block you are unstacking is clear.\newline
6. Once you pick up or unstack a block, you are holding the block.\newline
7. You can only put down a block that you are holding.\newline
8. You can only stack a block on top and not under of another block if you are holding the block being stacked.\newline
9. You can only stack a block on top and not under of another block if the block onto which you are stacking the block is clear.\newline   
10. Once you put down or stack a block, your hand becomes empty.\newline
11. Once you stack a block on top of a second block, the second block is no longer clear.\newline\newline
Thinking like Alan Turing starting from the <Initial State> build a plan to get to the <Goal State>. For each action step carefully check that the step follows the rules. <Goal State> : Your goal is to have that the red block on top of the blue block and the orange block on top of the red block.\newline\newline
output format for each step until you reach the goal state:\newline
<state> : <state>\newline
<action> : < action to be performed in this step >\newline
<assess the action> : < are we building the stack bottom up, check carefully>\newline
\noindent
\end{large}
\end{tcolorbox}
\caption{The Zero-shot prompt used with the persona of Alan Turing and Action Rationale approach for solving the Planning - Blocksworld Domain Problem.}
\label{fig:planning_baseline_prompt_alan_turing_rationale}
\end{figure*}

\begin{figure*}[!th]
\begin{tcolorbox}[
    enhanced,
    opacityback=0.2,
    opacityframe=0.2,
    colback=lightgray,
    colframe=lightgray, 
    top=1mm, 
    bottom=1mm, 
    left=1mm, 
    right=1mm,
    %sharp corners,
    fontupper=\sffamily\small,
    ]    
\begin{large}
\noindent
You are playing with a set of blocks where you need to arrange the blocks into stacks.\newline\newline
<Initial State> : As initial conditions you have that, the orange block is clear, the hand is empty, the blue block is on top of the red block, the orange block is on top of the blue block and the red block is on the table.\newline
<Goal State> : Your goal is to have that the red block on top of the blue block and the orange block on top of the red block.\newline\newline
Here are the actions you can do:\newline
-Pick up a block from the table\newline
-Unstack a block from on top of another block\newline
-Put down a block on the table\newline
-Stack a block on top of another block\newline
\newline
Rules:\newline
1. You can only pick up or unstack one block at a time.\newline
2. You can only pick up or unstack a block if your hand is empty.\newline
3. You can only pick up a block if the block is on the table and the block is clear. A block is clear if the block has no other blocks on top of it and if the block is not picked up.\newline
4. You can only unstack a block from on top of another block if the block you are unstacking was really on top of the other block.\newline
5. You can only unstack a block from on top of another block if the block you are unstacking is clear.\newline
6. Once you pick up or unstack a block, you are holding the block.\newline
7. You can only put down a block that you are holding.\newline
8. You can only stack a block on top and not under of another block if you are holding the block being stacked.\newline
9. You can only stack a block on top and not under of another block if the block onto which you are stacking the block is clear.\newline   
10. Once you put down or stack a block, your hand becomes empty.\newline
11. Once you stack a block on top of a second block, the second block is no longer clear.\newline\newline
Thinking like Alan Turing, starting from the <Initial State> build a plan to get to the <Goal State> . For each action step carefully check that the step follows the rules. Divide the task into smaller steps, starting with placing the bottom block first, followed by the middle blocks, and finally the top block. <Goal State> : Your goal is to have that the red block on top of the blue block and the orange block on top of the red block.\newline\newline
[PLAN]
\noindent
\end{large}
\end{tcolorbox}
\caption{The Zero-shot prompt used in the Alan Turing + Progressive Block Placement Approach for solving the Planning - Blocksworld Domain Problem.}
\label{fig:planning_baseline_prompt_alan_turing_progressive}
\end{figure*}

\begin{figure*}[!th]
\begin{tcolorbox}[
    enhanced,
    opacityback=0.2,
    opacityframe=0.2,
    colback=lightgray,
    colframe=lightgray, 
    top=1mm, 
    bottom=1mm, 
    left=1mm, 
    right=1mm,
    %sharp corners,
    fontupper=\sffamily\small,
    ]    
\begin{large}
\noindent
As a math professor, you will solve the given problem accurately '<using algebra>'. If there is no exact match choose the closest option.\newline

Question: John found that the average of 15 numbers is 40. If 10 is added to each number then the mean of the numbers\newline
Answer Choices: (A) 50 (B) 45 (C) 65 (D) 78 (E) 64\newline
Approach: <using algebra>\newline
Let's denote the sum of the 15 numbers as S. Since the average of the 15 numbers is 40, we can write the equation:
S / 15 = 40
S = 40 * 15 = 600
Now, we add 10 to each number, so the new sum is S + 15 * 10 = 600 + 150 = 750. The new average is:
New average = 750 / 15 = 50\newline
The answer is: (A) 50\newline

Question: If a / b = 3/4 and 8a + 5b = 22,then find the value of a.  \newline
Answer Choices: (A) 1/2 (B) 3/2 (C) 5/2 (D) 4/2 (E) 7/2\newline
Approach: <using algebra>\newline
To solve the given problem, we can use substitution. Since a / b = 3/4, we can write that as:
a = 3b/4
Now, substitute this expression for a into the second equation:
8(3b/4) + 5b = 22
Simplify and solve for b:
6b + 5b = 22
11b = 22
b = 2
Now that we have the value of b, we can find the value of a:
a = 3b/4
a = 3(2)/4
a = 6/4
a = 3/2
So, the value of a is (B) 3/2.\newline
The answer is: (B) 3/2\newline

Question: A person is traveling at 20 km/hr and reached his destiny in 2.5 hr then find the distance?\newline
Answer Choices: (a) 53 km (b) 55 km (c) 52 km (d) 60 km (e) 50 km\newline
Approach: <using algebra>\newline
Using the formula distance = speed × time, we can calculate the distance as follows:
Distance = 20 km/hr × 2.5 hr = 50 km
So, the closest answer choice is (e) 50 km.\newline
The answer is: (e) 50 km\newline\newline
Question: How many keystrokes are needed to type the numbers from 1 to 500?\newline  
Answer Choices: (a) 1156 (b) 1392 (c) 1480 (d) 1562 (e) 1788\newline
Approach: <using algebra>\newline
Let's break down the number of keystrokes needed into groups based on the number of digits:
One-digit numbers (1-9): There are 9 one-digit numbers, so we need 9 keystrokes.
Two-digit numbers (10-99): There are 90 two-digit numbers, each requiring 2 keystrokes, so we need 90 * 2 = 180 keystrokes.
Three-digit numbers (100-500): There are 401 three-digit numbers (500 - 100 + 1), each requiring 3 keystrokes, so we need 401 * 3 = 1203 keystrokes.
Now let's add up the keystrokes from all groups: 9 + 180 + 1203 = 1392.
Therefore, the closest option is (b) 1392.\newline
The answer is: (b) 1392\newline
\noindent
\end{large}
\end{tcolorbox}
\caption{AQuA few-shot-CoT prompt using algebraic approach.}
\label{fig:aqua_few_shot_prompt_algebra_approach}
\end{figure*}

\begin{figure*}[!th]
\begin{tcolorbox}[
    enhanced,
    opacityback=0.2,
    opacityframe=0.2,
    colback=lightgray,
    colframe=lightgray, 
    top=1mm, 
    bottom=1mm, 
    left=1mm, 
    right=1mm,
    %sharp corners,
    fontupper=\sffamily\small,
    ]    
\begin{large}
\noindent
Think like Alan Turing and solve the problem.If there is no exact match choose the closest option.\newline\newline
Question: John found that the average of 15 numbers is 40. If 10 is added to each number then the mean of the numbers\newline
Answer Choices: (A) 50 (B) 45 (C) 65 (D) 78 (E) 64\newline
Approach : <think like Alan Turing>\newline
Let's analyze the problem step by step:
The average of 15 numbers is 40.
So, the sum of these 15 numbers is 15 * 40 = 600.
10 is added to each number.
Now, since there are 15 numbers, the total increase in the sum will be 15 * 10 = 150.
The new sum of these 15 numbers will be 600 + 150 = 750.
To find the new mean, divide the new sum by the total number of values, which is 15.
New mean = 750 / 15 = 50.
Therefore, the closest option to the new mean is (A) 50.\newline
The answer is: (A) 50\newline\newline
Question: If a / b = 3/4 and 8a + 5b = 22,then find the value of a.\newline
Answer Choices: (A) 1/2 (B) 3/2 (C) 5/2 (D) 4/2 (E) 7/2\newline
Approach : <think like Alan Turing>\newline
Let us analyze the given information and solve the problem step by step.
We are given that a / b = 3/4, which can be written as a = (3/4) * b.
We are also given the equation 8a + 5b = 22.
Substitute the value of a from step 1 into the equation from step 2:
8 * (3/4) * b + 5b = 22
Simplify the equation:
6b + 5b = 22
Combine the terms and solve for b:
11b = 22
b = 2
Now that we have the value of b, let's find the value of a using the equation from step 1:
a = (3/4) * b
a = (3/4) * 2
Simplify to find the value of a:
a = 3/2
Thus, the value of a is 3/2, which corresponds to answer choice (B).\newline
The answer is: (B)\newline\newline
Question: A person is traveling at 20 km/hr and reached his destiny in 2.5 hr then find the distance?\newline  
Answer Choices: (a) 53 km (b) 55 km (c) 52 km (d) 60 km (e) 50 km  \newline
Approach : <think like Alan Turing>\newline
To find the distance, we can use the formula: distance = speed × time.
In this case, the speed is 20 km/hr and the time is 2.5 hours.
Distance = 20 km/hr × 2.5 hr = 50 km
So, the closest option is (e) 50 km.\newline
The answer is: (e) 50 km\newline\newline
Question: How many keystrokes are needed to type the numbers from 1 to 500?  \newline
Answer Choices: (a) 1156 (b) 1392 (c) 1480 (d) 1562 (e) 1788\newline
Approach : <think like Alan Turing>\newline
To determine the number of keystrokes needed to type the numbers from 1 to 500, let's break it down by digit count:
One-digit numbers (1-9): There are 9 one-digit numbers, each requiring 1 keystroke. Total: 9 keystrokes.
Two-digit numbers (10-99): There are 90 two-digit numbers, each requiring 2 keystrokes. Total: 90 * 2 = 180 keystrokes.
Three-digit numbers (100-500): There are 401 three-digit numbers, each requiring 3 keystrokes. Total: 401 * 3 = 1203 keystrokes.
Adding the keystrokes for each group, we get: 9 + 180 + 1203 = 1392 keystrokes.
Thus, the closest option is (b) 1392\newline
The answer is: (b) 1392
\noindent
\end{large}
\end{tcolorbox}
\caption{AQuA few-shot-CoT prompt with the persona of Alan Turing.}
\label{fig:aqua_few_shot_prompt_algebra_approach_alan_turing}
\end{figure*}

\begin{figure*}[!th]
\begin{tcolorbox}[
    enhanced,
    opacityback=0.2,
    opacityframe=0.2,
    colback=lightgray,
    colframe=lightgray, 
    top=1mm, 
    bottom=1mm, 
    left=1mm, 
    right=1mm,
    %sharp corners,
    fontupper=\sffamily\small,
    ]    
\begin{large}
\noindent
Think like Dr. Patel, the mathematician who specializes in algebra and solve the problem. If there is no exact match choose the closest option.\newline\newline
Question: John found that the average of 15 numbers is 40. If 10 is added to each number then the mean of the numbers\newline
Answer Choices: (A) 50 (B) 45 (C) 65 (D) 78 (E) 64\newline
Approach : <think like Dr. Patel>\newline
Let's consider the problem step by step. John found that the average of 15 numbers is 40. To find the sum of these 15 numbers, we can use the formula:
Sum = Average × Number of elements
Sum = 40 × 15
Sum = 600
Now, 10 is added to each of the 15 numbers. So, we can find the new sum by adding 10 × 15 = 150 to the original sum:
New Sum = 600 + 150
New Sum = 750
To find the new average (mean), we can use the formula:
New Average = New Sum ÷ Number of elements
New Average = 750 ÷ 15
New Average = 50\newline
The answer is: (A) 50\newline\newline
Question: If a / b = 3/4 and 8a + 5b = 22,then find the value of a.\newline
Answer Choices: (A) 1/2 (B) 3/2 (C) 5/2 (D) 4/2 (E) 7/2\newline
Approach : <think like Dr. Patel>\newline
To solve this problem, first we can rewrite the equation a / b = 3/4 as a = (3/4)b.
Now we can substitute this expression for a into the second equation: 8a + 5b = 22.
Substitute a with (3/4)b:
8(3/4)b + 5b = 22
Simplify the equation:
6b + 5b = 22
Combine the terms:
11b = 22
Now, divide both sides by 11:
b = 2
Now that we have the value of b, we can find the value of a using the equation a = (3/4)b:
a = (3/4)(2)
a = 3/2
The value of a is 3/2. So, the correct answer is (B) 3/2.\newline
The answer is: (B) 3/2\newline\newline
Question: A person is traveling at 20 km/hr and reached his destiny in 2.5 hr then find the distance?  \newline
Answer Choices: (a) 53 km (b) 55 km (c) 52 km (d) 60 km (e) 50 km  \newline
Approach : <think like Dr. Patel>\newline
To find the distance, we can use the formula:
Distance = Speed × Time
Here, the speed is given as 20 km/hr and the time taken is 2.5 hours. So, we can plug these values into the formula:
Distance = 20 km/hr × 2.5 hr
Distance = 50 km\newline
The answer is: (e) 50 km\newline\newline
Question: How many keystrokes are needed to type the numbers from 1 to 500? \newline 
Answer Choices: (a) 1156 (b) 1392 (c) 1480 (d) 1562 (e) 1788\newline
Approach : <think like Dr. Patel>\newline
To determine the number of keystrokes needed to type the numbers from 1 to 500, let's break down the problem by analyzing the number of digits in each range:
For numbers 1-9, there are 9 numbers with 1 digit each, requiring 9 keystrokes.
For numbers 10-99, there are 90 numbers with 2 digits each, requiring 90 * 2 = 180 keystrokes.
For numbers 100-499, there are 400 numbers with 3 digits each, requiring 400 * 3 = 1200 keystrokes.
For the number 500, it has 3 digits, requiring 3 keystrokes.
Now, we can sum up the keystrokes for each range:
9 (for 1-9) + 180 (for 10-99) + 1200 (for 100-499) + 3 (for 500) = 1392 keystrokes.\newline
The answer is: (b) 1392
\noindent
\end{large}
\end{tcolorbox}
\caption{AQuA few-shot-CoT prompt with the persona of Dr. Patel: A renowned mathematician}
\label{fig:aqua_few_shot_prompt_algebra_approach_dr_patel}
\end{figure*}

\begin{figure*}[!th]
\begin{tcolorbox}[
    enhanced,
    opacityback=0.2,
    opacityframe=0.2,
    colback=lightgray,
    colframe=lightgray, 
    top=1mm, 
    bottom=1mm, 
    left=1mm, 
    right=1mm,
    %sharp corners,
    fontupper=\sffamily\small,
    ]    
\begin{large}
\noindent
As a math professor, use 3 distinct approaches and without using built-in algorithms, write python programs to  color the following graph, described as a set of edges, such that no two vertices on the same edge share a color. \newline \newline 
You may use at most 3 colors. \newline
Vertex 0 is connected to vertex 7. \newline
Vertex 0 is connected to vertex 8. \newline
Vertex 0 is connected to vertex 9. \newline 
Vertex 0 is connected to vertex 11. \newline 
Vertex 1 is connected to vertex 13. \newline 
Vertex 2 is connected to vertex 9. \newline 
Vertex 3 is connected to vertex 8. \newline 
Vertex 3 is connected to vertex 11. \newline 
Vertex 3 is connected to vertex 12. \newline 
Vertex 4 is connected to vertex 12. \newline 
Vertex 5 is connected to vertex 11. \newline  
Vertex 6 is connected to vertex 9. \newline 
Vertex 7 is connected to vertex 10. \newline 
Vertex 7 is connected to vertex 13. \newline 
Vertex 9 is connected to vertex 11. \newline 
Vertex 10 is connected to vertex 13. \newline 
Vertex 11 is connected to vertex 13. \newline 
There are a total of 14 vertices. Please label every vertex, even if it is disconnected from the rest of the graph.Please provide each vertex's color. Do not skip any vertices. Each color must be provided on a new line in the response and should be formatted as "{VERTEX NUMBER}: {VERTEX COLOR ASSIGNMENT (Color n)}". \newline  \newline 
Output format: \newline  
Approach 1 <name of the approach> : < python program from scratch to color the given graph accurately > \newline Approach 2 <name of the approach> : < python program from scratch to color the given graph accurately> \newline 
Approach 3 <name of the approach> : < python program from scratch to color the given graph accurately>
\noindent
\end{large}
\end{tcolorbox}
\caption{Graph Coloring prompt using a programming approach in the zero-shot setting.}
\label{fig:graph_coloring}
\end{figure*}

\end{document}